\documentclass{article}

\usepackage[nonatbib,preprint]{neurips_2022}

\usepackage{graphicx}
\usepackage{amsmath}
\usepackage{amssymb}
\usepackage{booktabs}
\usepackage{tabularx}
\usepackage{algorithm}
\usepackage{algpseudocode}
\usepackage[roman]{parnotes}
\usepackage{tablefootnote}

\usepackage[pagebackref,breaklinks,colorlinks]{hyperref}
\newcommand{\our}{MAPLE-X}
\newcommand{\AN}{\mathbb{A}}
\newcommand{\dataset}{NASBench-X}

\usepackage[capitalize]{cleveref}
\crefname{section}{Sec.}{Secs.}
\Crefname{section}{Section}{Sections}
\Crefname{table}{Table}{Tables}
\crefname{table}{Tab.}{Tabs.}

\begin{document}

\title{MAPLE-X: Latency Prediction with Explicit Microprocessor Prior Knowledge}
\author{Saad Abbasi\\
Waterloo AI Institute\\
University of Waterloo\\
Waterloo, Ontario, Canada\\
\and
Alexander Wong\\
Waterloo AI Institute\\
University of Waterloo\\
DarwinAI\\
Waterloo, Ontario, Canada\\
\and
Mohammad Javad Shafiee\\
Waterloo AI Institute\\
University of Waterloo\\
DarwinAI\\
Waterloo, Ontario, Canada\\
}
\maketitle

\begin{abstract}
    Deep neural network (DNN) latency characterization is a time-consuming process and adds significant cost to Neural Architecture Search (NAS) processes when searching for efficient convolutional neural networks for embedded vision applications. DNN Latency is a hardware dependent metric and requires direct measurement or inference on target hardware. A recently introduced latency estimation technique known as MAPLE predicts DNN execution time on previously unseen hardware devices by using hardware performance counters. Leveraging these hardware counters in the form of an implicit prior, MAPLE achieves state-of-the-art performance in latency prediction. Here, we propose \our{} which extends MAPLE by incorporating explicit prior knowledge of hardware devices and DNN architecture latency to better account for model stability and robustness. First, by identifying DNN architectures that exhibit a similar latency to each other, we can generate multiple virtual examples to significantly improve the accuracy over MAPLE. Secondly, the hardware specifications are used to determine the similarity between training and test hardware to emphasize training samples captured from comparable devices (domains) and encourages improved domain alignment. Experimental results using a convolution neural network NAS benchmark across different types of devices, including an Intel processor that is now used for embedded vision applications, demonstrate a 5\% improvement over MAPLE and 9\% over HELP. Furthermore, we include ablation studies to independently assess the benefits of virtual examples and hardware-based sample importance.
\end{abstract}

\vspace{-0.25in}
\section{Introduction}
\label{sec:intro}
Designing deep neural network (DNN) architectures, especially for efficient convolutional neural networks for embedded applications, is a tedious process, requiring multiple iterations involving  exploration of different layer connectivity choices and configurations. This motivated the deep learning community to transition from manually designed DNN architectures to automatic discovery of optimal DNNs via Neural Architecture Search (NAS) algorithms~\cite{tan2019mnasnet,wu2019fbnet}.

A NAS algorithm evaluates a wide variety of DNN architectures by employing a searching strategy and selects optimal models by assessing metrics such as model accuracy, efficiency and energy consumption. However, measuring architecture latency is laborious and time consuming. This is because in practice each DNN must be deployed on the physical target device to measure latency and energy characteristics~\cite{tan2019mnasnet}. However, this approach is difficult to scale due to the large number of architectures that a NAS evaluates. As an example, ProxylessNAS~\cite{cai2018proxylessnas} evaluates 300,000 architectures in its first-round. Assuming the average latency is 50 ms and 50 measurements are acquired to reduce variance, ProxylessNAS would need 750 hours -- equating to approximately a month of continuous data collection. This engineering effort is further exacerbated if multiple hardware devices must be targeted.

\begin{figure}[t!]
\vspace{-0.3in}
    \centering
    \includegraphics[width=0.85\columnwidth]{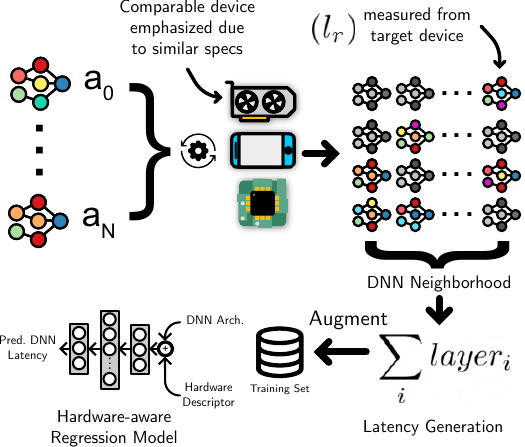}
    \caption{MAPLE-X overview. DNN architectures exhibiting a latency within certain bounds are clustered into neighborhoods. Given a DNN neighborhood, a layer-wise predictor is used to generate virtual examples which augment the training set. The virtual examples enable the hardware-aware regression model to adapt to the target device.}
    \label{fig:overview}
    \vspace{-0.3in}
\end{figure}

A variety of techniques have been proposed to predict DNN architecture latency in lieu of measuring them. Recently, layer-wise predictors have been employed to estimate architecture latency by summing up the execution time of each DNN layer~\cite{wu2019fbnet}. However, the layer-wise latency predictors are only effective on simple hardware which sequentially execute DNN layers. A layer-wise predictor typically fails to yield an accurate latency estimate on complex hardware as it does not capture the complexities of parallel or out-of-order execution. More recently, neural network based regression models have been introduced to predict latency on the target hardware. However, these methods require building architecture latency datasets for each target device which quickly becomes difficult to scale due to the wide variety of available hardware~\cite{dudziak2020brp}. Layer-wise or regression-based predictors therefore, generally pose a trade-off between estimator development time and prediction accuracy. To address this issue,  more recent methods aim to adapt regression models to the target hardware~\cite{lee2021help} instead of developing a model for every device. As such, meta-learning techniques have been used to adapt an already trained latency estimator to new devices. These methods employ a pool of training devices from which architecture latency has already been collected. The trained model is then adapted to the target device by measuring a small set of architecture latencies. Thus, a model adaptation approach significantly reduces the time it takes to adapt to unseen hardware.

\begin{figure}
\vspace{-1 cm}
\begin{tabularx}{\linewidth}{cc}
\setlength{\tabcolsep}{0.01cm} 
     \includegraphics[width=0.45\linewidth]{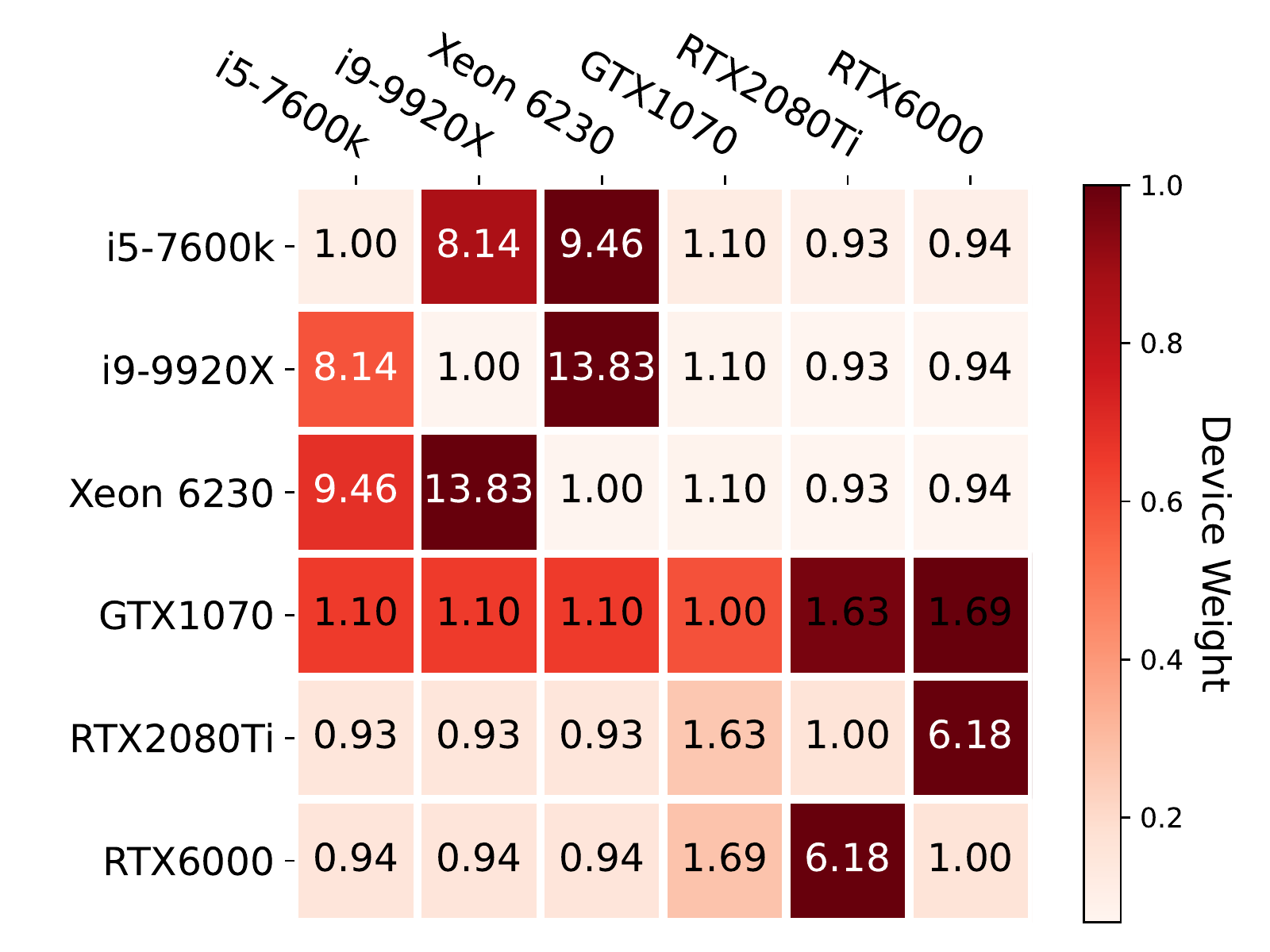} &  \includegraphics[width=0.45\linewidth]{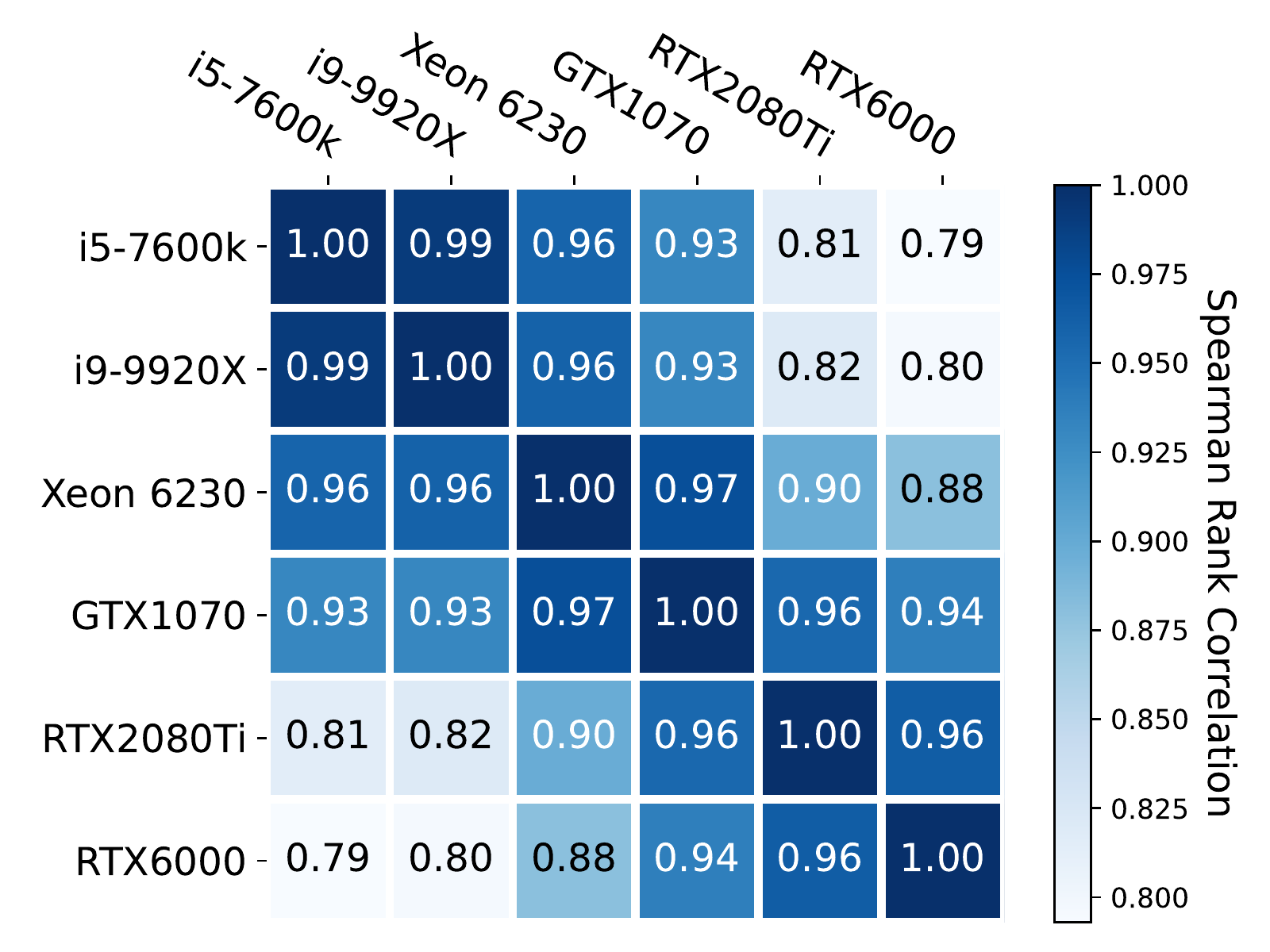}\\
\end{tabularx}
    \caption{\textbf{(left)} Device weights based on hardware specifications. Each device in the row can be considered as a test-device while the corresponding columns can be considered as the training device pool. \textbf{(right)} Spearman rank correlations of DNN architecture latency across all devices used in this study. From the figure, we note a higher correlation within device classes (CPUs or GPUs). Additionally, the rank is also tightly correlated within same micro-architectures. For instance, both Turing architecture GPUs (RTX2080Ti and RTX6000) exhibit a higher rank correlation with each other than the GTX1070, which features the Pascal architecture.}
\label{fig:rank}
\end{figure}

A recently introduced technique known as MAPLE~\cite{abbasi2021maple} employs hardware performance counters to characterize how different devices behave under specific workloads. MAPLE subsequently measures a small number of samples (i.e., 3 to 10) from the target hardware and augments the training set prior to optimizing a hardware-aware regression model. This augmentation enables rapid adaptation to the target hardware. While the hardware characterization and subsequent adaptation samples allow MAPLE to achieve state-of-the-art results, MAPLE leverages the prior hardware knowledge implicitly that the training pool may provide and thus might be leaving some performance on the table. Leveraging prior knowledge in machine learning has been shown to improve performance, particularly in cases where training samples are limited. In this study, our goal is to extend MAPLE by incorporating explicit prior knowledge provided by the training pool through two key mechanisms. First, we incorporate the architecture latency from the training devices as prior knowledge to generate virtual examples which improve the adaptation accuracy of the latency estimator. Second, we introduce a sample weighting scheme that emphasizes samples captured from devices similar to the target device. For example, if the test device is a 4000 CUDA-core GPU, a higher importance weight is assigned to samples captured from similar core-count GPUs compared to samples from CPUs or less similar GPUs. The proposed weighing scheme leads to a more accurate adaptation of the latency predictor by adapting the model via taking advantage of more similar domain.

\section{\our{} Design}

In this section we describe the proposed extensions to MAPLE and the architecture latency dataset. \our{} incorporates the prior knowledge present in the training pool samples to generate virtual examples and, additionally, emphasizes samples from devices similar to the target hardware.

\subsection{Hardware-based Importance Sampling}
Conventional MAPLE places significantly higher emphasis on samples acquired from the target device. The rationale behind this design choice was that higher adaptation weights would allow the latency predictor to adapt to the target device with higher accuracy. In contrast, here we place more emphasis on samples acquired from devices that are similar to the target device. This makes intuitive sense as devices comparable to the target device would exhibit similar latency characteristics. For instance, if we assume that the target device is an Nvidia RTX6000 GPU, the knowledge gained from a RTX2080Ti is significantly more relevant than any CPU. We use publicly available device specifications such as core-count, maximum clock-speed and the thermal design power to compute the distance between the target device and the training devices. Since the device specification descriptor is only three-dimensional, a euclidean distance can be used to compute the distance between the target and the training devices. The weights are subsequently computed by $\frac{1}{\sqrt{d_i}}$ where $d_i$ is the distance between the target device and the device $i$ in the training set. The square root ensures the weight magnitudes do not differ from each other by a large degree, which may cause instability during training.

\subsection{DNN Latency Neighborhood Prior}
In this section, we provide evidence of the knowledge present in the training pool and elaborate upon the mechanism through which it is integrated. The integration is dependent on five related ideas that we now explain.

\begin{figure}
\begin{tabularx}{\linewidth}{ccc}
\setlength{\tabcolsep}{0.01cm} 
     \includegraphics[width=0.31\linewidth]{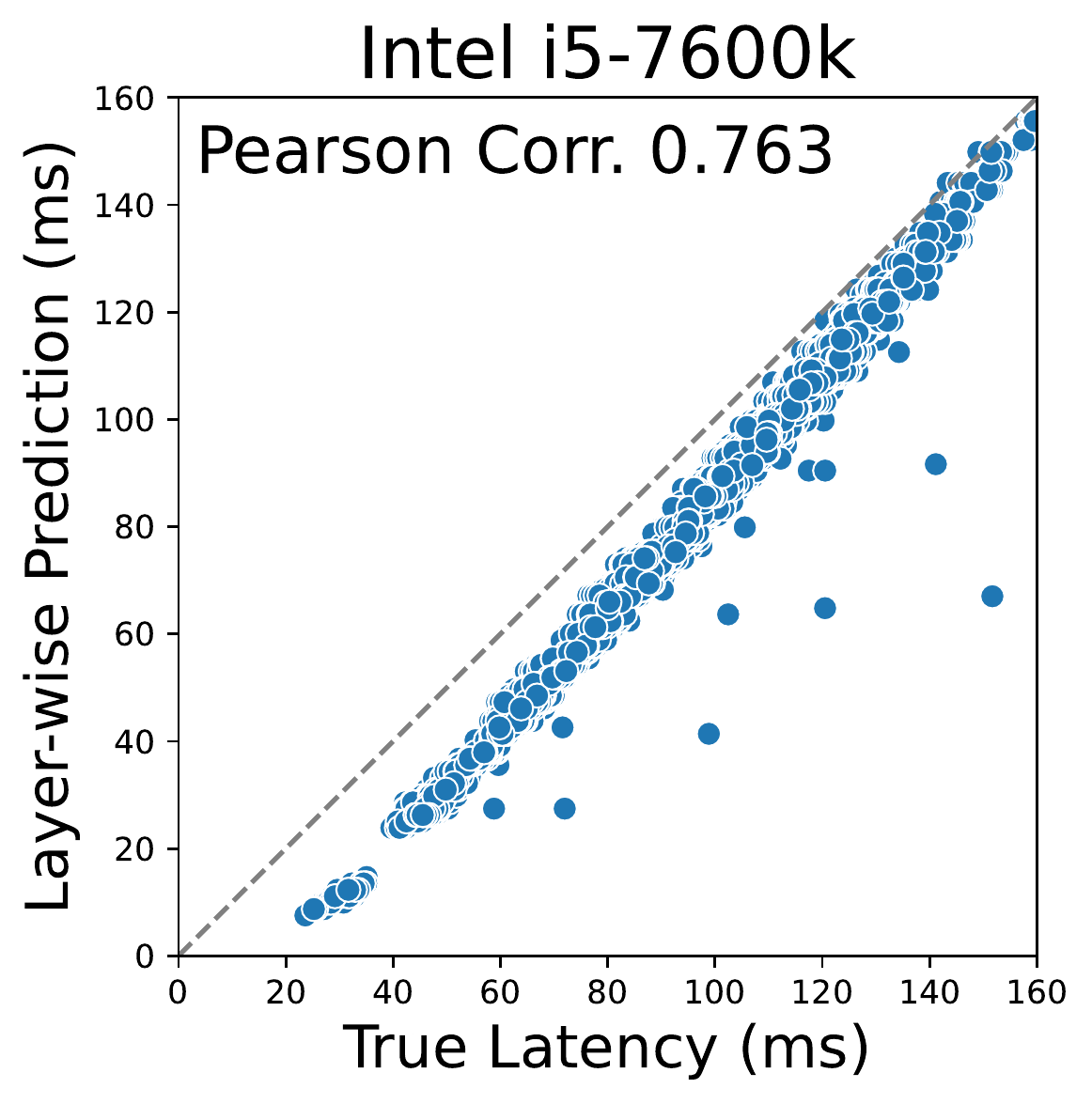} &  \includegraphics[width=0.31\linewidth]{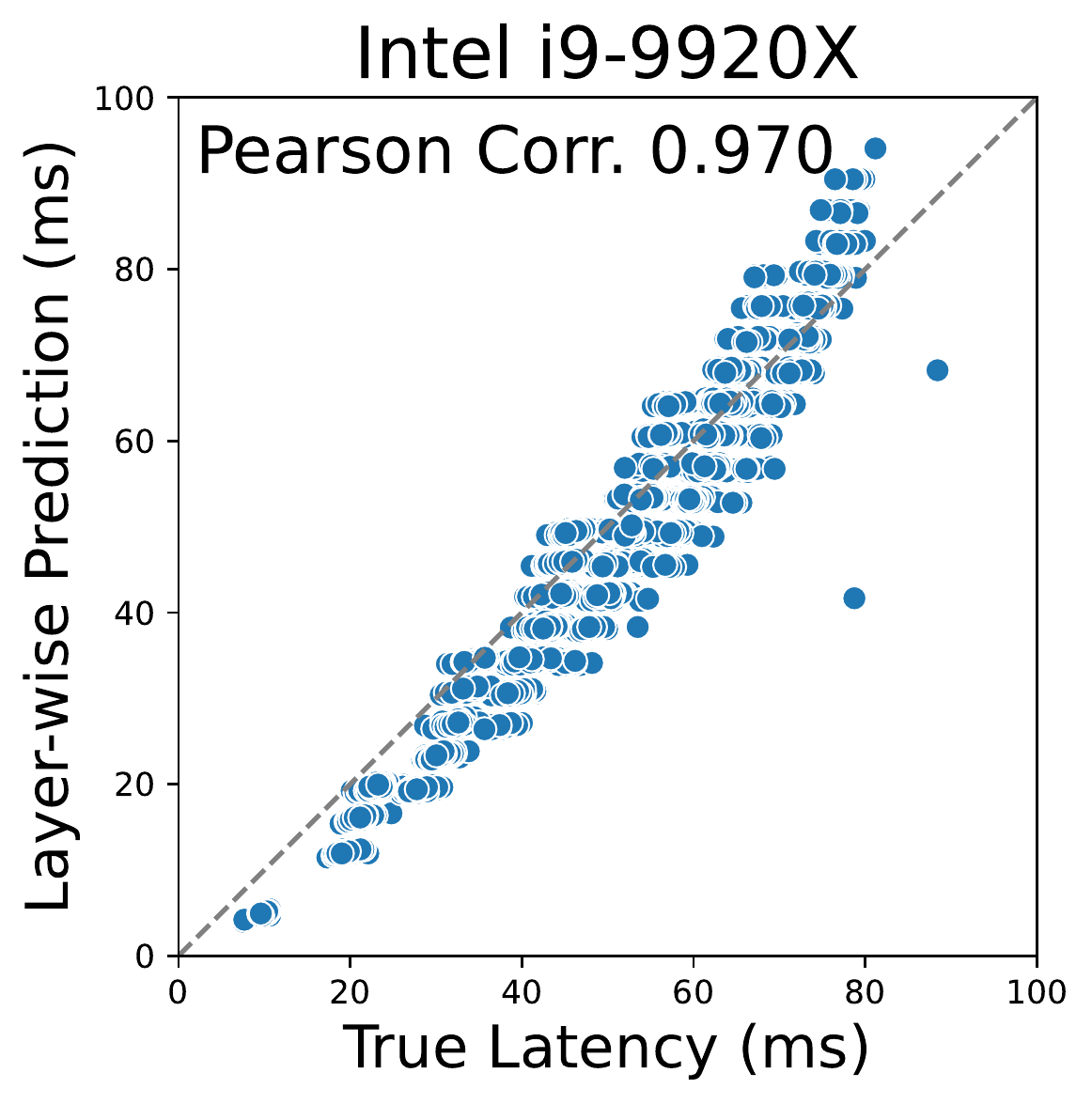} &
     \includegraphics[width=0.30\linewidth]{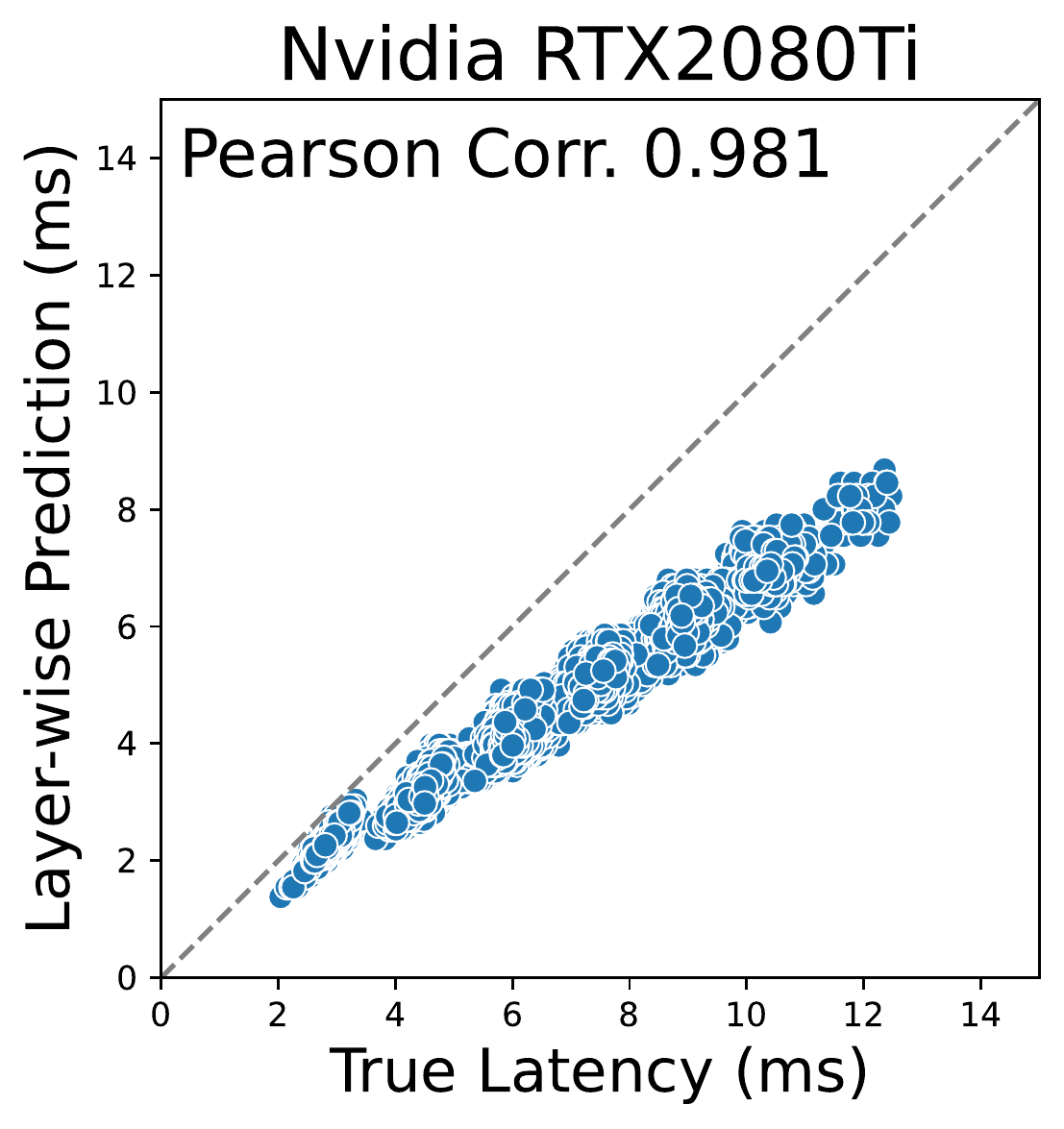}
\end{tabularx}
\caption{Correlation between the true latency and layer-wise summed latency for three devices. Despite a high prediction error for Intel i5-7600k (used for embedded vision applications) and Nvidia RTX2080Ti, a strong Pearson correlation enables MAPLE-X to use layer-wise estimation as a proxy for relative latency.}
\label{fig:corr}
\end{figure}

\vspace{-0.35cm}
\paragraph{Latency Ranking.} DNN latency is a function of its architecture as well as the underlying hardware. While in general, all DNN architectures would exhibit a lower latency on a faster device, the optimal architecture may differ across devices as different hardware architectures may prioritize different DNN operations. However, we observe a high latency rank correlation within comparable hardware micro-architectures (Figure~\ref{fig:rank}). A high rank correlation implies that DNN architectures exhibit a consistent ordering on comparable devices. As a consequence of this, architectures that are close in latency on one device would also exhibit similar latencies on a comparable device. Leveraging this high rank correlation between devices, it is possible to identify DNN architectures that consistency exhibit a similar latency to each other across different hardware.
\vspace{-0.35cm}
\paragraph{Neighborhood Discovery.}We identify DNN neighborhoods by finding architectures that exhibit a latency within certain bounds from a given architecture ($a_r$). For each given architecture, we denote such neighborhoods (sets) as $\AN_r = \big\{ a_{0}, a_{1} \ldots	 a_{n} \big\}$ where $a_{n}$ is a neighbor architecture for $a_r$ defined by the neighborhood $\AN_r$. By identifying such latency neighborhoods across training devices, we can infer which architectures are likely to be close in latency on the target hardware. Importantly, the neighborhood prior cannot typify what the true latency of the samples in $\AN_r$ would be on the target device, but only a belief that certain architectures would exhibit latencies within certain bounds. Therefore, we select an architecture ($a_{r}$) from a neighborhood and measure its latency ($l_{r}$) on the target device. 
\vspace{-0.35cm}
\paragraph{Prior Strength.} 
A DNN neighborhood represents a set of architectures that exhibit similar latency. Architectures in $\AN$ can simply be assigned a random latency centered on $l_r$ between some bounds. However, this approach leads to a weak prior as it would not take into account how the architecture latency is distributed on the target hardware. More concretely, while we know $l_{a0} \approx l_{a1}$ in set $\AN_r$, a stronger prior can be developed if it is possible to incorporate the relative latencies of samples in $\AN_r$ (i.e. by what proportion does $l_{a0}$ exhibit a higher or lower latency than $l_{a1}$).

\vspace{-0.35cm}
\paragraph{Latency Modeling.}
Layer-wise estimators typically exhibit poor accuracy on complex hardware. Fortunately, the layer-wise summed latency still exhibits a strong correlation with the true latency (Figure~\ref{fig:corr}). The cause of this correlation is better illustrated with an example. Consider $a_i$ consists of two convolution operations and three average pools. If the architecture in $a_j$ consists of three convolutions, it would exhibit a higher latency than $a_i$. Although a layer-wise estimator would not be able predict the absolute latency accurately, it would still be able to encapsulate the relative increase in latency as this increase is primarily due to the addition of an operation.

\begin{figure}
\begin{tabularx}{\linewidth}{ccc}
\setlength{\tabcolsep}{0.01cm} 
     \includegraphics[width=0.31\linewidth]{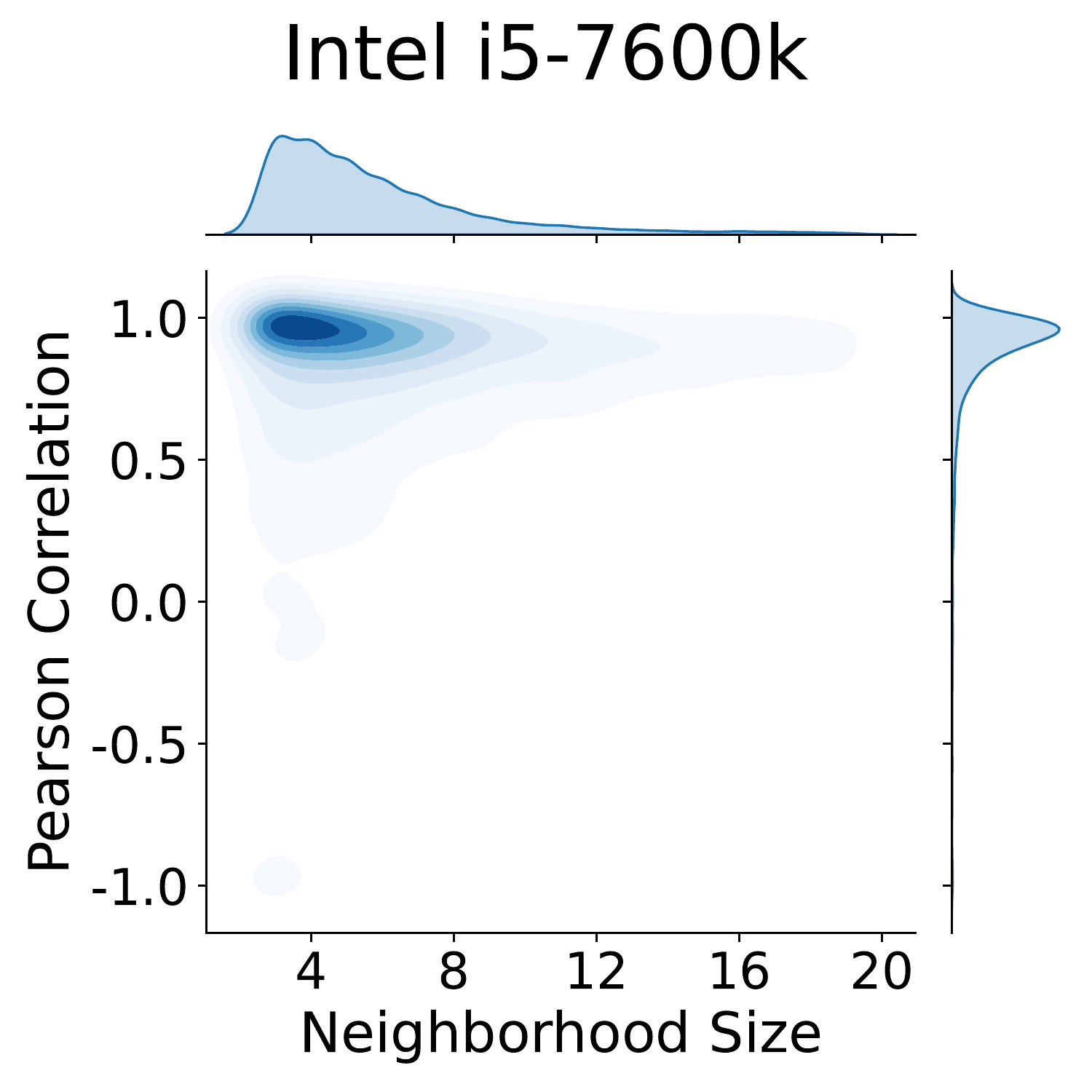} &  \includegraphics[width=0.31\linewidth]{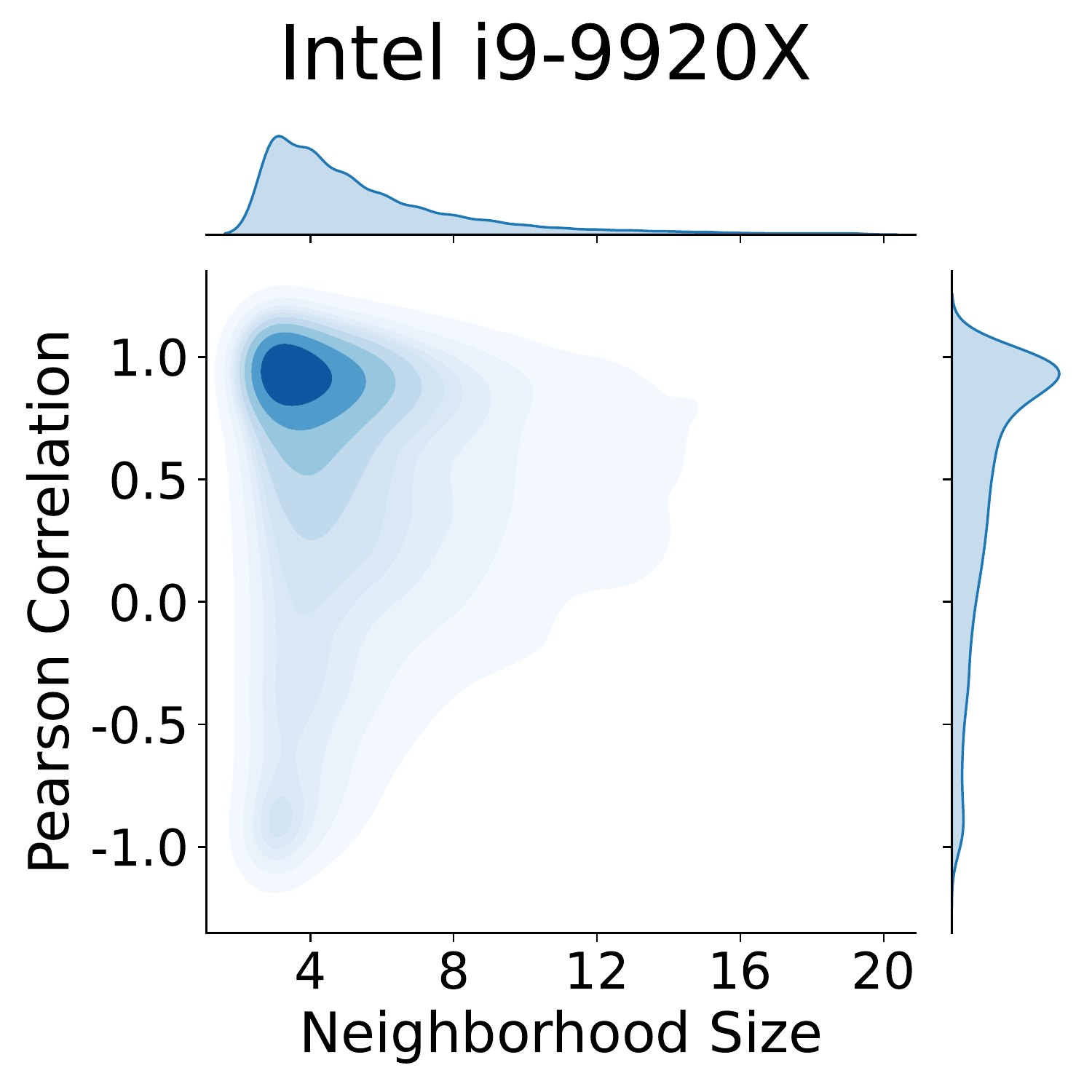} &
     \includegraphics[width=0.30\linewidth]{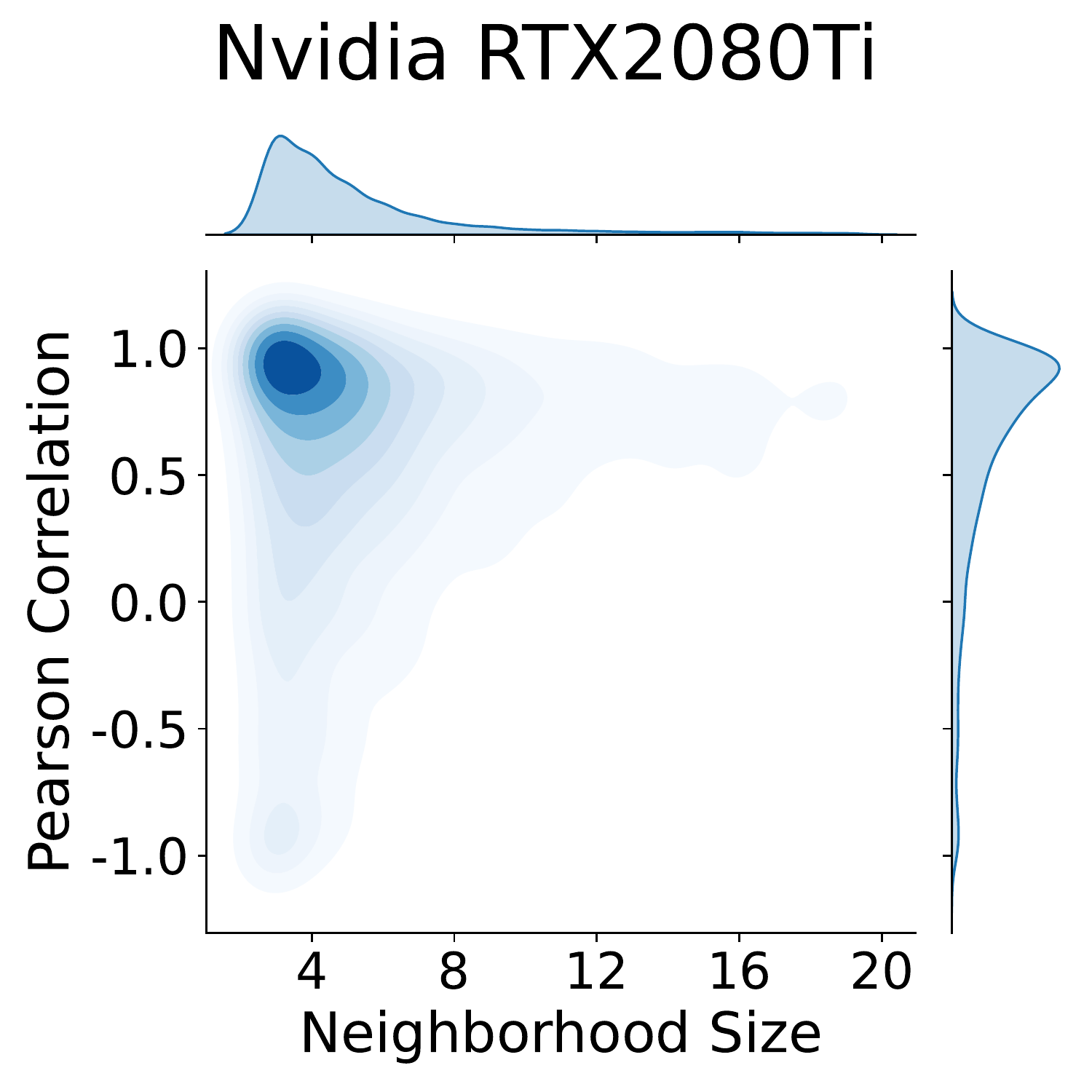}
\end{tabularx}
    \caption{Pearson correlation density plots. The correlation between the layer-wise derived latency and true latency is computed for each DNN neighborhood in the prior. The correlations are denser between a neighborhood size of 2-10, with the majority of neighborhoods exhibiting a correlation above 0.5. Thus, we can use LUT-derived latency from the target device as a proxy for the relative latency of DNN architectures in a neighborhood.}
\label{fig:density}
\end{figure}

Given that there is a strong linear correlation between the layer-wise latency and true execution time, we now determine if this correlation holds for DNN neighborhoods of varying size. We compute Pearson correlation between the layer-wise and the true latency for every DNN neighborhood. We subsequently plot a density map (Figure~\ref{fig:density}) between the latency correlation and the size of DNN neighborhoods to assess if most neighborhoods exhibit a strong correlation. From Figure~\ref{fig:density}, we note most DNN neighborhoods typically exhibit correlation values between 0.7--1.0 for sizes between 2--10, making it probable that any neighborhood we sample from that area could have its relatively latency accurately modelled by a layer-wise predictor.

\vspace{-0.35cm}
\paragraph{Latency Label Generation.}
Encouraged by this observation and considering MAPLE already characterizes each operation present in the architectural search space, we construct a layer-wise predictor to estimate the relative latency of architectures in a given neighborhood. Using the layer-wise predictor, we compute the layer-wise latency of each architecture in $\AN$. Subsequently, we subtract the mean layer-wise latency of $\AN$ from each DNN architecture. This leaves only the relative differences intact. We then add $l_{r}$ as an offset to each architecture in $\AN_r$ which yields our virtual latency labels for the architectures in $\AN$. More formally, we can compute the virtual latency labels with the following expression: $v_{a_n} = LUT(a_n) - \mu_{t} + l_{r}$ where $LUT(\cdot)$ outputs the layer-wise derived latency for a given architecture and $\mu_{t}$ is the mean latency calculated by $\frac{\sum_{a_i \in \AN} LUT(a_i)}{\| \AN\| }$.
Together, $\AN$ and $V_\AN = \{v_{a_0},v_{a_1}, \ldots, v_{a_n}\}$, represent virtual examples generated based on an explicit measurement $l_{r}$ given the prior knowledge $\AN$ and $LUT(\cdot)$. These virtual examples, along with explicitly measured samples, are mixed in with the dataset prior to training. The number of virtual examples generated depends upon the size of the neighborhood selected. 

\vspace{-0.1in}
\section{Experiment \& Discussion}

\subsection{Evaluation}
\vspace{-0.1in}
\label{sec:eval}
To evaluate our methodology, we use the same dataset as MAPLE which we refer to as \dataset{}~\cite{abbasi2021maple}. \dataset{} is based on  NASBench-201, a cell-based convolutional neural network NAS dataset and includes a total of 15,625 architectures. Extending NASBench-201, \dataset{} provides latency measurements of all architectures present in NASBench-201 on six different devices including Intel i5-7600k (a processor that is now used for embedded vision applications), i9-9920X, Xeon-6230, Nvidia GTX1070, RTX2080Ti and RTX6000. In addition, \dataset{} also includes measurements from ten different hardware counters measuring number of cpu-cycles, instructions, cache accesses, among others. We follow a similar methodology as MAPLE and use a regression model for latency prediction. Keeping true to the MAPLE framework, we encode DNN architectures via one-shot encoding and concatenate the hardware performance counters to yield a descriptor for each device. Finally, to compare our results we use 10\% error-bound accuracy as our primary metric. This metric  calculates the proportion of predictions that falls within $\pm10\%$ of the true latency.

\begin{table}
\centering
\begin{tabular}{lrrrr} 
\toprule
&\multicolumn{4}{c}{$\pm$ 10\% Error-bound Accuracy}\\
Test Device & \our{} & \our{}1 & MAPLE & HELP \\ 
\hline
i5-7600k    & 0.92    & 0.91 & 0.85 & 0.93 \\
i9-9920X    & 0.90    & 0.84 & 0.94 & 0.93  \\
Xeon 6230   & 0.90    & 0.90 & 0.88 & 0.77  \\
GTX1070     & 0.84    & 0.85 & 0.90 & 0.95  \\
RTX2080Ti   & 0.98    & 0.88 & 0.75 & 0.75  \\
RTX6000     & 0.87    & 0.83 & 0.78 & 0.53  \\ 
\hline
Mean        & 0.90    & 0.87 & 0.85 & 0.81  \\
\bottomrule
\end{tabular}
\vspace{0.5cm}
\caption{Comparison of prediction accuracy between {\our}, MAPLE and HELP. MAPLE-X1 represents an ablation study and demonstrates improvement with just the DNN neighborhood prior (no Hardware-based importance sampling). MAPLE and \our{} use three adaptation samples while HELP uses ten, as recommended by the original study.}
\label{tab:results}
\end{table}

\subsection{Results}
In this section we evaluate and compare \our{} with MAPLE~\cite{abbasi2021maple} and HELP~\cite{lee2021help}. We adopt a one-device-leave-out approach and form a training pool of five devices listed in Section~\ref{sec:eval} and use the sixth device for testing. We sample the same 900 architectures from each device to yield a total of 4500 training samples. Since we want to demonstrate the advantage of including virtual examples, we restrict ourselves to just three adaptation points from the test device. The adaptation samples yield three architecture neighborhoods with varying sizes between 5-10 samples per neighborhood which are mixed in with the training set.

Table~\ref{tab:results} compares our results with MAPLE and HELP. We note that incorporating prior knowledge from training devices leads to a mean 5\% improvement over MAPLE and 9\% over HELP. Integrating the prior knowledge in the form of virtual examples improves performance due to two key reasons. First, virtual examples represent additional samples from the target device which aid the adaptation of the hardware-aware regression model. Second, since MAPLE only adapts with just 3-10 samples, most batches do not contain an adaptation sample. The inclusion of virtual examples ensures that most batches during model training include a sample representative of the target device. This, in turn, enables the optimizer to move in a direction on the loss landscape more amicable to the target device at each gradient update. By emphasizing samples from more comparable devices, we allow the model to learn more effectively.

To decouple the improvement due to solely building a DNN neighborhood prior, we include results of an ablation study in Table~\ref{tab:results} with the so-called \our{}1. We observe that solely including the DNN neighborhood prior improves results over MAPLE by an average of 2\% and illustrates the effectiveness of Hardware-based importance sampling in improving the accuracy by another 3\%. 

{\small
\bibliographystyle{ieee_fullname}
\bibliography{egbib}
}

\end{document}